\documentclass[12pt]{article}

\usepackage{rotating}

\usepackage{tikz}
\usepackage[utf8]{inputenc}
\usepackage[T1]{fontenc}
\usepackage{amsmath}
\usetikzlibrary{fit,positioning}
\usepackage[]{algorithm2e}
\usepackage{booktabs,tabu}
\usepackage{bm}
\usepackage{array}
\usepackage{enumerate}
\usepackage{url}
\usepackage{amsfonts}
\newcolumntype{C}{@{\extracolsep{1cm}}c@{\extracolsep{2pt}}}%

\newcommand{\Othlang}{\ell\!\!\backslash}

\newcommand{\BiLDA}{\texttt{biLDA}}

\newcommand{\LDA}{\texttt{LDA}}
\newcommand{\segBiLDA}{\texttt{segBiLDA}}
\newcommand{\muLDA}{\texttt{biLDA}}

\allowdisplaybreaks
\begin{document}

\title{\Large Bilingual Topic Models for Comparable Corpora }

\author{\large Georgios Balikas$^\star$, Massih-Reza Amini$^\dagger$, Marianne Clausel$^\ddagger$\\
$^\star$ Salesforce, Meylan France\\
$^\dagger$ University of Grenoble Alpes, France\\
$^\ddagger$ University of Lorraine, France}

\date{}
\maketitle

\begin{abstract}

Probabilistic topic models like Latent Dirichlet Allocation (LDA) have been previously extended to the bilingual setting.
A fundamental modeling  assumption in several of these extensions is that the input corpora are in the form of document pairs whose constituent documents share a single topic distribution.
However, this assumption is strong for comparable corpora that consist of documents thematically similar to an extent only, which are, in turn, the most commonly available or easy to obtain.
In this paper we relax this assumption by proposing for the paired documents to have separate, yet bound topic distributions. 
We suggest that the strength of the bound should depend on each pair's semantic similarity.
To estimate the similarity of documents that are written in different languages we use cross-lingual word embeddings that are learned with shallow neural networks. 
We evaluate the proposed binding mechanism by extending two topic models: a bilingual adaptation of LDA that assumes bag-of-words inputs and a model that incorporates part of the text structure in the form of boundaries of semantically coherent segments.  
To assess the performance of the novel topic models we conduct intrinsic and extrinsic experiments on five bilingual, comparable corpora  of English documents  with  French, German, Italian, Spanish and Portuguese documents. The results demonstrate the efficiency of our approach in terms of both topic coherence measured by the normalized point-wise mutual information, and generalization performance measured by perplexity and in terms of Mean Reciprocal Rank in a cross-lingual document retrieval task for each of the language pairs.
\end{abstract}

\section{Introduction}

Extracting information from the ever growing amount of documents available in more than one language is one of the challenges of Natural Language Processing. Any important progress on this domain has a significant impact on many applications ranging from speech recognition and intelligent interfaces to machine translation.

\begin{figure}[t]\centering
 \includegraphics[scale=0.9]{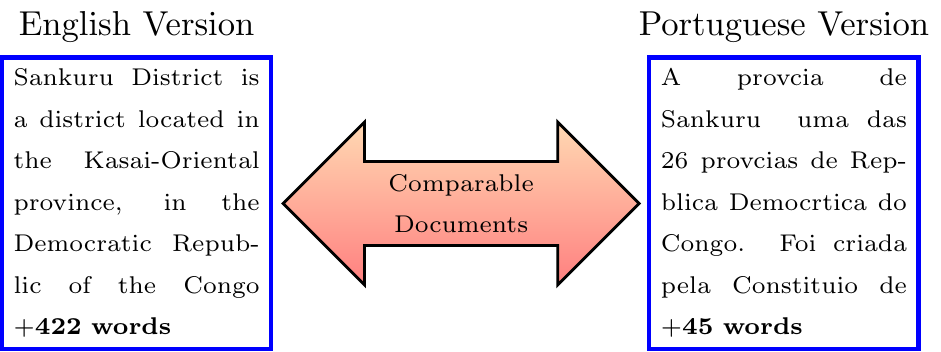}
 \caption{Motivating Example: excerpts from comparable Wikipedia documents. The English version is several times bigger than the Portuguese and one may reasonably assume it covers more topics.}\label{fig:motivatingExample}
\end{figure}

\begin{figure}[t]\centering
 \includegraphics[width=\textwidth]{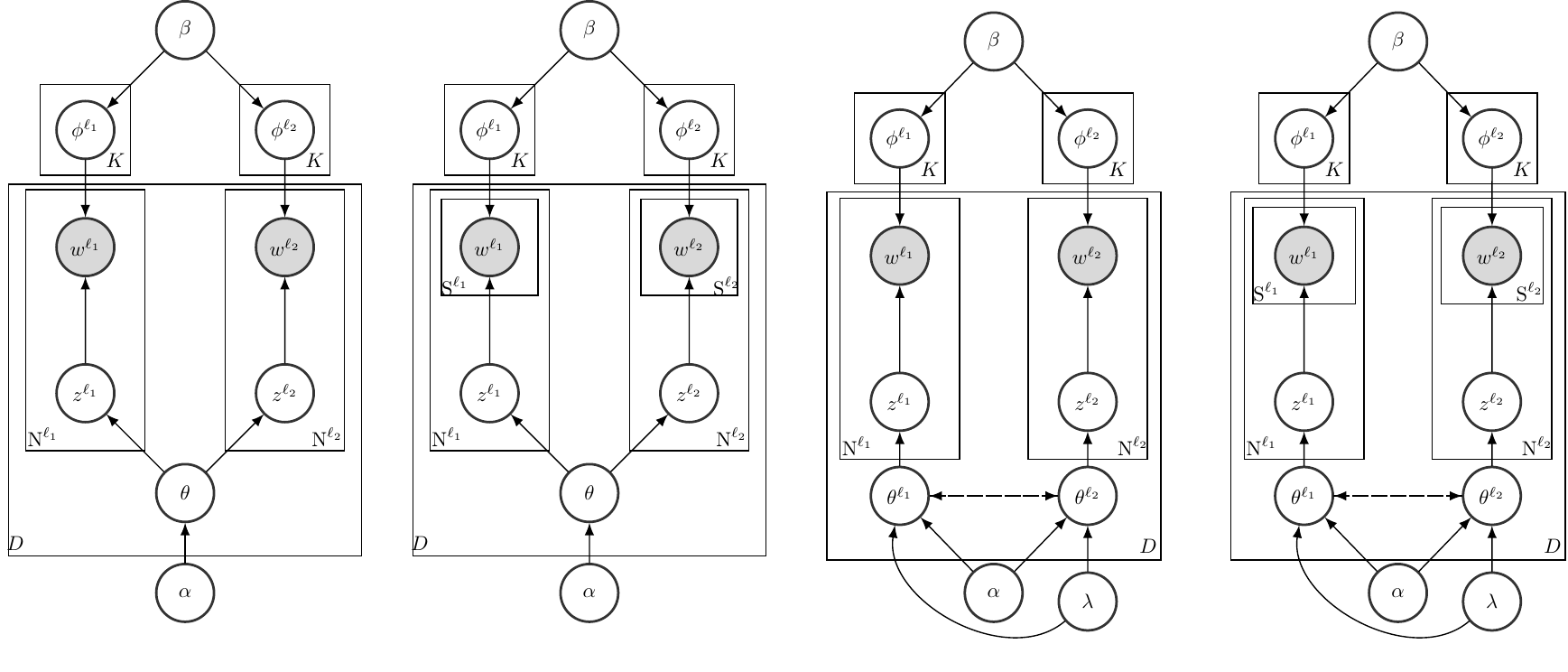}
 \caption{The topic models used in this work. From left to right: (i) {\BiLDA} (ii)  {\segBiLDA} (iii)  $\lambda$-{\BiLDA} (iv) $\lambda$-{\segBiLDA}. The difference of {\BiLDA} and {\segBiLDA} from their $\lambda-$ counterparts lies on the fact that the second have separate but bound topic distributions and the strength of binding is controlled by  $\lambda$. }\label{fig:topic_models}
\end{figure}

Probabilistic topic models  like Latent Dirichlet Allocation (LDA) \cite{blei2003latent} are a family of unsupervised models that when applied to monolingual collections uncover the latent themes underlying it. Following their success in the monolingual setting, they have been extended to the bilingual setting. 
The most representative bilingual topic model is illustrated in Figure \ref{fig:topic_models}(i) and is commonly called bilingual\footnote{Depending on the number of input languages the model may be referred to as either bilingual or multilingual LDA.} LDA (\muLDA) \cite{mimno2009polylingual,de2011knowledge,vulic2015probabilistic}. It extends LDA and does not require any prior, language dependent  linguistic knowledge but the input collection to be in the form of pairs of thematically aligned documents. Given the pairs, a fundamental hypothesis of {\muLDA} is that the documents of a pair share a single  per-document topic distribution $\theta$. This entails that the documents in a  pair discuss exactly the same themes. Although reasonable for parallel corpora, whose pairs consist of documents that are translations, this assumption is strong for collections composed by pairs of \textit{comparable} documents (e.g. \cite{McEnery2007}), that is documents similar to some extent only. Figure \ref{fig:motivatingExample} illustrates an example of comparable documents written in English and Portuguese.\footnote{This is a real example from the Wiki$_{\text{En-Pt}}$ collection used in our experiments.} As the English document is larger, one would expect it to cover more topics. Hence, having a shared topic distribution between those two documents is a strong assumption.

In this paper we propose to extend bilingual topic models by relaxing the assumption of comparable documents sharing a single  topic distribution. 
For this purpose, instead of a shared distribution we allow the documents of a pair to have two, separate, yet \textit{bound} distributions.
We suggest that the strength of the bound should depend on the semantic similarity of the documents of the pair. The estimation of this similarity for documents  written in different languages is a task on itself.  
Instead of using dictionaries, which are one-to-one or one-to-$N$ discrete word associations and do not capture different levels of similarity,
or machine translation systems, which are computationally expensive to develop,
we propose to use cross-lingual word embeddings. Word embeddings are continuous vector representations of words that encode their semantic properties by projecting similar words close in a vector space \cite{harris1954distributional,abs-1301-3781}. The cross-lingual embeddings, complementary to projecting similar words close in the semantic space for each language, project similar words across languages close in this space. Their  potential  has been previously shown for various tasks such as cross-lingual information retrieval \cite{vulic2015monolingual}.

The questions  we attempt to answer are twofold: $(Q1)$ How to better adapt bilingual topic models to comparable collections? $(Q2)$ Does this adaptation generalize well across different types of topic models?
To address these questions, the paper proposes three contributions: 
\begin{enumerate}[(i.)]
\item 
a novel approach that combines topic models  with (shallow) neural networks for learning word embeddings allowing the former to extract latent distributions from comparable corpora,
\item the extension of  {\muLDA}  and of a monolingual topic model that incorporates text structure in the form of boundaries of coherent text spans, 
\item a systematic evaluation of the  novel topic models on five comparable corpora where English are paired with  French, German, Italian, Spanish and Portuguese documents. \end{enumerate}

We evaluate the topic models both on \textit{intrinsic} and \textit{extrinsic} tasks. Our results demonstrate the efficiency of our approach in terms of topic coherence measured by the normalized point-wise mutual information of the induced topics, generalization performance measured by calculating the perplexity of held-out documents, and in terms of Mean Reciprocal Rank in a cross-lingual document retrieval task for each of the language pairs. In the rest of the paper we provide a detailed discussion of these points.

\section{Related Work}

Our work lies in the intersection of the fields of multilingual topic modeling and  cross-lingual word embeddings. We review  the relevant literature starting with the work on multilingual topic models. 

\paragraph{Multilingual Topic Models} There are two different lines of research in the multilingual topic modeling approaches with respect to the nature of the available training inputs. The first line  assumes access or attempts to create linguistic resources such as dictionaries, in order to identify the topical links and alignments between the multilingual documents of a text corpus
\cite{boyd2009multilingual,jagarlamudi2010extracting,zhang2010cross,boyd2010holistic}. The topic alignments between documents are not implicit in the input, and the models identify the topically relevant multilingual documents and the topic distributions while leveraging the available linguistic resources.
For instance, \cite{boyd2009multilingual} propose the multilingual topic model for unaligned text (MuTo) that discovers a parallelism in the documents of the corpus at the vocabulary level, while it assumes that similar themes are expressed in both languages. To perform the joint task of producing consistent topics in each of the two languages and then aligning them, the model uses dictionaries. \textit{JointLDA} is a model with similar motivations, proposed in \cite{jagarlamudi2010extracting}. To cope with the multilingual setting, \textit{jointLDA} also uses dictionaries but learns topics shared among the input languages. Those topics are distributions over the vocabulary terms of the multilingual corpus, and as a result, terms of different languages occur in a topic. Despite the advantages of such models, their requirements for several language-specific resources can be seen as  a limitation.

The second, more flexible line of research, investigates topic modeling solely on the basis of the textual inputs. Those inputs, usually consist of text corpora with documents that are either parallel translations \cite{zhao2006bitam} or comparable translations of each other \cite{ni2009mining,mimno2009polylingual,de2009cross,platt2010translingual}. Such topic models by not relying on any  external resource are a better fit for unsupervised methods. The most representative  model of this family is {\BiLDA}, which extends {\LDA} in the bilingual \cite{ni2009mining,mimno2009polylingual,de2009cross} or the multilingual setting \cite{krstovski2013online,mimno2009polylingual}. The difference between the bi- or multilingual setting lies on the number of the input languages, which ranges from two to several. 

The model we propose in this work belongs to this family of models as it assumes access to a corpus whose input documents form theme-aligned pairs. However, our model is more flexible as instead of assuming a single topic distribution per pair of documents, it uses two topic distributions that are linked with a binding mechanism that uses cross-lingual word embeddings to account for the level of similarity between the documents. 


\bigskip

\textbf{Cross-lingual Word Embeddings} 
According to the \textit{distributional hypothesis} \cite{firth1957synopsis,harris1954distributional,PessiotKAG10}, linguistic items such as words with similar distributions should have similar meanings. In other words, semantically  similar  words  should have  similar  contextual distributions. The contextual information is usually induced assuming the context to be documents or sliding windows and is represented by populating  word-context co-occurrence matrices. 
For words, different models that learn distributed representations have been recently proposed and those models are used as implementation models of the distributional hypothesis \cite{turian2010word}. 
To this end, the \textit{distributed representations}  (also known as word embeddings) associate words with dense vectors, of dimension of a few hundreds to some thousands.
A distributed representation of a symbol is a vector of features that characterize the meaning of the symbol and in our case a symbol is a word. The representation is a continuous $D$-dimensional vector and, therefore, compact in the sense that an exponential number of symbols (words) in the number of dimensions can be efficiently represented \cite{turian2010word}, compared, for instance, to the one-hot-encoding scheme that can only represent $D$ symbols when using $D$ dimensions.

Among different models, the skipgram   model with negative sampling \cite{abs-1301-3781},  has been shown to be efficient and effective in several applications \cite{Amini07}.
Such a model is a function $f$ that projects a word $w$ in a $D$-dimensional space: $f(w) \in \mathbb{R}^D$, where $D$ is predefined. 
Although the model relies on a well-defined prediction task \cite{baroni2014don}, it has been shown that  
it is implicitly factorizing
a word-context matrix, whose cells are the pointwise mutual information (\textit{PMI}) of
the respective word and context pairs, shifted by a global constant \cite{LevyG14,LiXTJZC15}.
Despite the theoretical equivalence however, an advantage of the skipgram model compared to other models that factorize such matrices like Latent Semantic Indexing \cite{hofmann1999probabilistic} is its ability to practically scale to huge amounts of data.

The skipgram model was initially proposed for the single language setting. However, motivated by the idea of having a single representation space shared by more languages, 
cross-lingual word embeddings models extended the idea in the bilingual and multilingual settings. 
The models can be grouped with regard to the approach used to align the cross-lingual embeddings. Models like \cite{mikolov2013exploiting} followed by \cite{xing2015normalized,lazaridou2015hubness} for instance, 
begin by learning monolingual word embeddings and try to learn a linear transformation from one space to the other. Another way to learn cross-lingual embeddings  \cite{gouws2015simple,duong2016learning,vulic2016bilingual} is by artificially generating  multilingual documents by concatenating the documents of parallel or comparable corpora and then training existing monolingual models. Lastly, models like  \cite{luong2015bilingual,GouwsBC15} perform  joint optimization of monolingual and cross-lingual losses. They can benefit from very big monolingual corpora for optimizing their monolingual objectives while relying on smaller corpora for optimizing their cross-lingual objective. 

In this work we use Bilbowa \cite{GouwsBC15}. It belongs in the family of models that jointly optimize  monolingual and cross-lingual objectives. It extends the skipgram model for cross-lingual  embeddings and trains directly on monolingual
data. It uses a bilingual signal from a smaller
set of raw-text sentence-aligned data to align the cross-lingual embeddings. 

%
%

\bigskip

\textbf{Combining Topic Models and Word Embeddings}
While topic models are trained to infer the per-word and per-document topic distributions, the skipgram model is trained by trying to predict the context of a word. Different efforts have attempted to extend the models by combining them. 
For instance,  embeddings associate words with a single vector, which may be limiting for encoding the different meanings of polysemous words. This limitation  motivated works that extend word embeddings with topic models. The purpose is for the topic models to uncover the  different senses of a word, so that different embeddings can be  derived for each sense \cite{LiuLCS15,ChengWWYC15}. Such efforts attempt to produce better performing word embeddings.

In a relevant line of work, the purpose is to produce better topic models while taking advantage of the fact that text embeddings model semantic similarity.
To this end, \cite{NguyenBDJ15,DasZD15,ZhangZ16} extend  topic models in order to  encourage
the models to group words that are \textit{a priori} known to be semantically related into topics, where the \textit{a priori} knowledge comes from training embeddings in large external corpora.

Our work belongs to this second line of research because we use word embeddings to improve the results of topic models. Differently from previous research though, word embeddings are used only to estimate the similarity of documents written in different languages. Also, our models are multilingual, while all previous work investigated the intersection of topic models and word embeddings in the monolingual setting using English.

\section{Framework}
Our primary goal in this work is to  adapt the bilingual topic models for comparable corpora. To accomplish that we relax the assumption of paired documents having identical topic distributions. In the rest of this section, after presenting the notation, we briefly discuss {\muLDA} in $\S$\ref{section:muLDA}.
To illustrate how several classes of topic models can benefit by the adaptation to comparable corpora, we introduce a novel bilingual topic model that incorporates parts of the document's structure ($\S$\ref{section:segBiLDA}). 
We, then, extend {\muLDA} and the novel bilingual topic model for comparable corpora in $\S$\ref{section:lMuLDA} and $\S$\ref{section:lsegBiLDA}. 

\medskip

\begin{table}[t]
\scriptsize\centering\renewcommand{\arraystretch}{1.2}

 \begin{tabu}{ll}\hline
  $\ell_1, \ell_2$ & Input languages e.g., $\ell_1$=English, $\ell_2$=German \\
  $d_i, d_i^{\ell_1}, d_i^{\ell_2}$ & doc. pair $d_i$, whose aligned docs. are $d_i^{\ell_1}$, $d_i^{\ell_2}$ \\
  $\lambda_i$ & The semantic similarity between $d_i^{\ell_1}$ and $d_i^{\ell_2}$ \\  \tabucline[0.4pt  off 3pt]{-}
  $
  V_{\ell}$ & The size of the vocabulary in language $\ell$\\
  $\Omega_{k,i}$ & The number of words in $d_i$ assigned to topic $k$ \\
  $\theta_i$ & Topic distribution of $d_i$ \\
  $\Psi_{k,w}$ & The number of assignments of word $w$ to topic $k$\\ 
  $\phi_k$ & Word distribution for topic $k$ \\
  $s_{i,j}$& The $j^{th}$ segment  of document $d_i$ \\
  $w_{i,j,k}$& The $k^{th}$ word of segment $s_{i,j}$   \\
  $N_i$ & The number of words of  document $d_i$\\
  $N_{i,j} $ & The number of words in segment $s_{i,j}$ \\
  $N_{i,j,w}$ & The number of occurrences of word $w$ in  $s_{i,j}$  \\
  \hline
 \end{tabu}
\caption{The notation used for the development of the topic models. Adding exponents $\ell_1, \ell_2$ to the symbols of the lower part of the table (below the dashed line) stands for counts specific to $d_i^{\ell_1}, d_i^{\ell_2}$. }\label{tbl:notation}
\end{table}

The notation is summarized in Table \ref{tbl:notation}. For consistency, we keep the symbols of previous work \cite{wang2008distributed} to the extent of possible. We denote $\ell_1$ and $\ell_2$ the different languages present in a comparable corpus. As languages are handled symmetrically, for convenience we designate by $\Othlang$, the language different from language $\ell\in\{\ell_1,\ell_2\}$.  
The inputs of the topic models are document pairs $d_i=(d_i^{\ell_1}, d_i^{\ell_2})$, that consist of thematically aligned documents $d_i^{\ell_1}$ and  $d_i^{\ell_2}$, written in $\ell_1$ and $\ell_2$.  Depending on the model, documents are either represented as a bag-of-words, or as a bag-of-segments. Segments are text spans smaller than documents, for instance sentences, and are represented as a bag-of-words. 
Considering $\ell\in\{\ell_1,\ell_2\}$, $s_{i,j}^{\ell}$ is the $j^{th}$ segment of document $d_i^{\ell}$. Segmented documents have a hierarchical structure:  they are composed by segments that are composed by words in turn. Depending on the model, there may exist either a single $\theta_i$ topic distribution that captures the topics present in both documents of the pair $d_i$, or two, separate yet \textit{bound} topic distributions $\theta_i^{\ell}$, $\theta_i^{\Othlang}$ that capture the topics of $d_i^{\ell}$ and $d_i^{\Othlang}$ respectively. The rest of the notation in Table \ref{tbl:notation} stands for count matrices or vectors used during inference.

\subsection{The bilingual LDA}\label{section:muLDA}

{\muLDA} (Figure \ref{fig:topic_models}(i)) is a direct adaptation of {\LDA} in the bilingual setting where a parallel collection is assumed to be the model's input. 
Due to its effectiveness we use it as a reference in this work.  
{\muLDA} assumes that the documents of an aligned pair $d_i$ have identical topic distributions (a single and shared $\theta_i$) and therefore discuss the same topics. 
Also, it expects the documents as a bag-of-words. Its generative story is as follows:

\medskip

\begin{small}
\begin{itemize}
  \item for each topic $k\in[1,K]$:
     $\phi_k^{\ell_1} \sim Dir(\beta)$, $\phi_k^{\ell_2} \sim Dir(\beta)$
  \item for each document pair $d_i$:
  \begin{itemize}
   \item sample $\theta_i \sim Dir(\alpha)$ 
    \item for each language $\ell\in\{\ell_1, \ell_2\}$
       \begin{itemize}

   \item for each of the $N^{\ell}_i$ words:
     \begin{itemize}
      \item sample $z \sim Mult(1, \theta_i)$
      \item sample $w \sim Mult(1, \phi^{\ell}_{z})$
     \end{itemize}
   \end{itemize}
  \end{itemize}
\end{itemize}
\end{small}

\medskip

\noindent The collapsed Gibbs sampling updates \cite{vulic2015probabilistic}  for the topic of  word $j$ of  document $d_i$ is $\forall \ell\in\{\ell_1,\ell_2\}$:
\scriptsize
\[
P\Big( z_{ij}^{\ell}=z_k|\boldsymbol{z}_{\neg{ij}}^{\ell}, \boldsymbol{z}^{\Othlang}, \boldsymbol{w}^{\ell}, \boldsymbol{w}^{\Othlang}, \alpha, \beta,  \Big) \propto  
\frac{\Psi_{k,w,\neg ij}^{\ell}+\beta}{\Psi_{k,\cdot,\neg ij}^{\ell}+V_{\ell}\beta}[\Omega_{i,k,\neg ij}+\alpha] \label{eq:lambdaInference}
\]
\normalsize
\noindent A dot ``$\cdot$'' occurring in the subscript of a count variable, stands for the summation over the possible values of the element it replaces,  i.e.,$\Psi_{k,\cdot,\neg ij}^\ell= \sum\limits_{w=1}^{V} \Psi_{k,w,\neg ij}^\ell$. Also, $\neg$ stands for excluding the counts of the particular variable with respect to a segment, e.g., $\neg ij$ excludes the counts of the $j$-th word of $d_i^\ell$. 
Further, $Dir(\alpha)$ stands for a sample from a Dirichlet distribution with prior $\alpha$ and $Mult(M, \theta)$ stands for $M$ samples from a Multinomial distribution parametrized by $\theta$. 

For {\muLDA}, as well as for the models we present next, we consider the Dirichlet hyperparameters $\alpha\in\mathbb{R}^K$ and  $\beta\in\mathbb{R}^{V}$ to have fixed values, implying symmetric priors. Extending the models to asymmetric priors or even learning their values could be done as in \cite{asuncion2009smoothing} for example. Also, as commonly done we omit from the generative stories the steps where the sizes of segments or documents are sampled as their sizes are observed during inference.
As noted, {\muLDA} uses a bag-of-words representation; next we present an extension that uses a more complex document representation. 

\subsection{Text structure incorporation }\label{section:segBiLDA}
In this section,  we propose \texttt{segment-BiLDA} ({\segBiLDA}) that incorporates prior knowledge of text structure using a more complex document representation than bag-of-words. 
Although important for inference, the bag-of-words assumption is limiting. 
In fact, previous research in the single language domain showed the benefits of similar extensions: Wang et al. \cite{wang2007topical}  proposed a model that handles bigrams as a single token or as two unigrams depending on the topic, Lau et al. \cite{lau2013collocations} modeled frequent bigrams as separate tokens,  Balikas et al. \cite{balikas2016on} proposed to incorporate sentence boundaries to LDA,  while Boyd et al. \cite{boyd2009syntactic} incorporated parse trees . These important contributions  focused on the monolingual setting and used English texts for empirical evaluation. Here, we extend topic models to account for text structure in the bilingual case.

For our subsequent analysis, we define coherent text segments to be contiguous words of a document that are topically coherent. 
A topically coherent text segment refers to a segment whose constituent words discuss a single or very few related themes. For instance, one would expect frequent bigrams like  ``information retrieval'' or even short sentences to be topically coherent as they generally convey a simple message. 
We model this property with ({\segBiLDA}), which is  illustrated in Figure \ref{fig:topic_models}(ii). {\segBiLDA} assumes that the input text is  segmented \textit{a priori} and incorporates the boundaries of segments in its generative story:

\medskip

\begin{small}
\begin{itemize}
 \item for each topic $k\in[1,K]$:
   $\phi_k^{\ell_1} \sim Dir(\beta)$, $\phi_k^{\ell_2} \sim Dir(\beta)$
 \item for each document pair $d_i$:
 \begin{itemize}
 
  \item sample $\theta_i \sim Dir(\alpha)$ 
  \item for each language $\ell\in\{\ell_1, \ell_2\}$
  \begin{itemize}
    \item for the $j$  segment ($1\le j \le S^{\ell}$ ):
    \begin{itemize}
    \item sample $z \sim Mult(1, \theta_i)$
    \item sample segment words: \\ $(w_1 \ldots w_{N_{i,j}^\ell}) \sim Mult(N^{\ell}_{i,j}, \phi^{\ell}_{z})$
    \end{itemize}
  \end{itemize}

 \end{itemize}
\end{itemize}
\end{small}

\medskip


The important difference of {\muLDA} from {\segBiLDA} (Figures \ref{fig:topic_models}(i) and \ref{fig:topic_models}(ii)) lies in the addition of the segment's plate. 
A topic is sampled per segment, and every word of a segment is associated with it. The segment boundaries limit the number of topics that appear in  the segment to be equal to one.  
As in {\muLDA} though, words remain the document units and this single topic is associated with each word of the segment.  Therefore, comparing the models on measures like perplexity that are calculated at the word level is fair.  
To infer these topics we propose a collapsed Gibbs sampling approach, that $\forall \ell\in\{\ell_1,\ell_2\}$, samples topics from:

\scriptsize\begin{align}
 ~& P(z_{s_{i,j}}^{\ell}=z_k|\boldsymbol{z}_{\neg s_{ij}}^{\ell}, \boldsymbol{z}^{\Othlang},  \boldsymbol{w}^{\ell}, \boldsymbol{w}^{\Othlang}, \alpha, \beta) \propto 
  [\Omega_{i,k, \neg s_{i,j}}+\alpha] \times \nonumber \\
 ~& \frac{\prod\limits_{w\in s_{ij}^{\ell}}(\Psi_{k,w,\neg s_{ij}}^{\ell}+\beta)\cdots(\Psi_{k,w,\neg s_{ij}}^{\ell}+\beta+(N_{i,j,w}^{\ell}-1))}{(\Psi_{k,\cdot,\neg s_{ij}}^{\ell}+\beta V_{\ell})\cdots(\Psi_{k,\cdot,\neg s_{ij}}^{\ell}+\beta V_{\ell}+(N_{i,j}^{\ell}-1))}
 \label{eq:finalSource}
\end{align}
\normalsize

For convenience, we discuss the derivation of Eq. \eqref{eq:finalSource} in the Appendix. 
In  Eq. \eqref{eq:finalSource}, the product appearing in the numerator of the second term results from the bag-of-words assumption for the words of segments. The (possibly multiple) occurrences of a word $w$ in a segment $s_{i,j}^{\ell}$, generated by the topic $k$, are taken into account by the factor $(\Psi_{k,w,\neg s_{ij}}^{\ell}+\beta)$, which is incremented by one for every other occurrence of the word after the first.  For example, if  word $w$ appears twice in $s_{i,j}^{\ell}$, then $N_{i,j,w}^{\ell}=2$, and the factor $(\Psi_{k,w,\neg s_{ij}}^{\ell}+\beta)(\Psi_{k,w,\neg s_{ij}}^{\ell}+\beta+1)$ denotes the contribution of the occurrences of the word to the probability that  $s_{i,j}^{\ell}$ is generated by the topic $k$. This way, every word of the segment contributes to the probability of sampling a particular topic. 
Similarly, the denominator acts as a normalization term. The progressive increase of its values can also be explained intuitively: given the bag-of-words  assumption of words within a segment, the product normalizes the probability of assigning the topic $k$ to a word of the segment, given that the previous words have also been assigned to this topic. Notice, that if the size of the segment is 1, the model as well as the sampling equations reduce to {\muLDA}.

The bag-of-words assumption in  {\muLDA} results in a joint distribution of random variables (here topics) being invariant to any permutation of the variables (exchangeability).  This  holds for {\segBiLDA} only locally, within segments. Globally, within a document, words are not exchangeable as the segment boundaries are utilized. While in {\muLDA} the topics of words are conditionally independent given the document's topic distribution, for {\segBiLDA} they are not, as they also depend on the rest of the segment's words.


Previous work in the monolingual case suggested to incorporate  various types of text structure to topic models ranging from $n$-grams to parse trees. 
{\segBiLDA} can be considered an extension of the model of Balikas et al. \cite{balikas2016on} in the bilingual setting. Rather than extending more complex models like Boyd's \cite{boyd2009syntactic} that may require parsing the documents, we opt for {\segBiLDA} due to the variety of segments it can handle. For instance, one can use linguistically motivated  segmentation approaches like sentence tokenization or statistically motivated segmentation approaches like frequent $n$-grams with the same model. Furthermore, these segmentation approaches can be accomplished efficiently and accurately across a wide range of languages without resorting to complex linguistic analysis tools like parsers. 


\subsection{Multilingual topic Extraction}\label{section:lMuLDA}
{\muLDA} and {\segBiLDA} assume a single topic distribution for the documents of a pair, which as illustrated in Figure \ref{fig:motivatingExample} is a string assumption for comparable documents. 
Apart from that, the motivations for adapting the  bilingual topic models to comparable corpora lie on two facts: on one hand, comparable corpora are more common and easy to obtain or to construct than parallel ones, which require additional linguistic knowledge and tools. On the other hand, recent advances on cross-lingual word embeddings resulted in methods that can be directly used to estimate the semantic similarity of documents written in different languages.  The latter facilitates the application of our method  to various pairs of languages without expensive resources.

For comparable corpora, we first propose the $\lambda$-{\BiLDA} model, whose graphical model is shown in Figure  \ref{fig:topic_models}(iii). 
In this case, instead of having a single, shared topic distribution we have a topic distribution per language shown as $\theta^{\ell_1}$ and $\theta^{\ell_2}$ in the figure. However, these distributions are bound between them. To model naturally the possible levels of dependence between $\theta^{\ell_1}$ and $\theta^{\ell_2}$ we need a binding mechanism flexible enough to model the two extreme conditions: total independence between the topic distributions of the aligned documents that should result in two distinct {\LDA} models (one per language), and a complete dependence between them (identical topic distributions)  which should result in {\BiLDA}. Similar dependence mechanisms were previously explored under the setting of streaming documents \cite{amoualian2016streaming}, where  topic distributions of earlier documents affect the distributions of later documents. 

The generative process for $\lambda$-{\BiLDA} is as follows:

\medskip

\begin{small}
\begin{itemize}
  \item for each topic $k\in[1,K]$: $\phi_k^{\ell_1} \sim Dir(\beta)$, $\phi_k^{\ell_2} \sim Dir(\beta)$
  \item for each document pair $d_i=(d_i^{\ell_1}, d_i^{\ell_2})$:
  \begin{itemize}
   \item sample $\lambda_i$ with respect to $d_i$
   \item sample $\theta_i^{\ell_1} \sim Dir(\alpha+ \lambda_i \theta_i^{\ell_2})$,   $\theta_i^{\ell_2} \sim Dir(\alpha+ \lambda_i  \theta_i^{\ell_1})$
    \item for language $\ell\in\{\ell_1, \ell_2\}$
       \begin{itemize}

   \item for each of the words $N^{\ell}$:
     \begin{itemize}
      \item sample $z \sim Mult(1, \theta_i^{\ell})$
      \item sample $w \sim Mult(1, \phi^{\ell}_{z})$
     \end{itemize}
   \end{itemize}
  \end{itemize}
\end{itemize}
\end{small}

\medskip



The central idea  is that the topic distributions  of documents in one language    depend on the topic distributions of documents in the other language via a binding mechanism that generates $\theta^{\ell}$ with a Dirichlet distribution depending on  $\theta^{\Othlang}; \theta^{\ell}|\theta^{\Othlang} \sim Dir(\alpha+ \lambda_i \theta_i^{\Othlang})$ and vice-versa. Note that from the Hammersley-Clifford theorem \cite{hammersley1971markov}, fixing the two conditional distributions $\theta^{\ell_1}|\theta^{\ell_2}$ and $\theta^{\ell_2}|\theta^{\ell_1}$ defines in an unique way the distribution of $(\theta^{\ell_1},\theta^{\ell_2})$ which implies that our generative process is well-defined. 

%
For inferring the topics of the observed words we propose a Gibbs sampling approach, whose derivation is given in the Appendix. The update equations for the topics of the words are then  $\forall \ell\in\{\ell_1,\ell_2\}$:

\begin{small}
\begin{align}
P\Big( z_{i,j}^{\ell}=z_k|\boldsymbol{z}_{\neg{i,j}}^{\ell}, \boldsymbol{w}^{\ell},\alpha, \beta, \lambda_i, \theta^{\Othlang} \Big) \propto \nonumber\\
\frac{\Psi_{k,w,\neg i,j}^{\ell}+\beta}{\Psi_{k,\cdot,\neg i,j}^{\ell}+V_{\ell}\beta}\cdot[\Omega_{i,k,\neg i,j}^{\ell}+\alpha+\lambda_i\theta_{d,k}^{\Othlang}] \label{eq:lambdaInference}
\end{align}
\end{small}

Gibbs sampling algorithms  obtain posterior samples by sweeping though each block of variables and sampling from their conditional, while the remaining blocks are fixed. 
In practice, the algorithm initializes randomly the topics of words. Then, during the Gibbs iterations and until convergence, sampling topics for words of $\ell$ assumes the distribution of $\theta^{\Othlang}$ fixed, and hence can be accessed as assumed by the generative story.

In Eq. \eqref{eq:lambdaInference}, $\lambda_i$ captures the dependency between the topic distributions of the documents of $d_i$. We use cross-lingual word embeddings for its estimation.  
We use the average (\texttt{avg}) compositional function of meaning, which was shown to be robust and effective \cite{MitchellL10,BlacoeL12}. Having the vectors of the document pair $d_i=(d_i^{\ell_1}, d_i^{\ell_2})$ in the embedded space, we then estimate $\lambda_i$ using the cosine similarity. Calculated in this way, $\lambda_i \in [-1,1]$ and since $\theta_{d,k}\in [0,1]$ it follows for  the second term of Eq. \ref{eq:lambdaInference} that $\Omega_{i,k,\neg i,j}^\ell >> \lambda_i\theta_{d,k}$, which results in negligible impact for $\lambda_i\theta_{d,k}$. To circumvent that we use:

\begin{scriptsize}
\begin{equation}                                                                                                                                                                                                                                                                                                                                                                                                                                                                                                                                                                                                                                                                                                                                                                                                                                                    
\lambda_i' = \lambda_i\times |N_i^{\Othlang}|=\Omega^{\Othlang}_{d,k}\label{eq:modifiedLambda}                                                                                                                                                                                                                                                                                                                                                                                                                                                                                                                                                                                                                                                                                                                                                                                                                                                     
\end{equation}
\end{scriptsize}

\noindent Notice that incorporating  Eq. \eqref{eq:modifiedLambda} to Eq. \eqref{eq:lambdaInference} has as an additional advantage the capacity to generalize previous models. In particular, it follows that {\muLDA} becomes a special case of $\lambda$-MuLDA with $\lambda=1$ (complete dependency where $\theta^{\ell}$ and $\theta^{\Othlang}$ are the same topic distributions). Also, when $\lambda=0$ (case of independence) we have two distinct {\LDA}s, one per language.\footnote{Although by definition $\lambda_i \in [-1,1]$ in all our experiments we found $\lambda_i$>0. }

\subsection{Combining the two models} \label{section:lsegBiLDA}
To this point, we proposed {\segBiLDA} that incorporates the boundaries of coherent segments like sentences, and $\lambda$-{\BiLDA}, that assumes bound topic distributions for the paired documents in the two languages.
The two models can be combined:  $\lambda$-{\segBiLDA} assigns consistent topics in the words of the segments of the documents and also assumes different topic distributions for each language. 

We illustrate   $\lambda$-{\segBiLDA} in Figure \ref{fig:topic_models}(iv). We omit the generative story, since it is a direct combination of the generative stories of {\segBiLDA} and $\lambda$-{\muLDA}. The inference process is given by the following equation, whose derivation is given in the Appendix. To sample the topics of segments from the conditional distribution for $\forall\ell\in\{\ell_1,\ell_2\}$: 

\begin{small}
\begin{align}\small
~& P\Big( z_{i,j}^{\ell}=z_k|\boldsymbol{z}_{\neg{i,j}}^{\ell}, \boldsymbol{w}^{\ell},\alpha, \beta, \lambda_i, \theta^{\Othlang} \Big) \propto
[\Omega_{i,k,\neg s_{i,j}}^{\ell}+\alpha+\lambda_i\Omega_{d,k}^{\Othlang}] \times\nonumber\\
~& \frac{\prod\limits_{w\in s_{ij}^{\ell}}(\Psi_{k,w,\neg s_{ij}}^{\ell}+\beta)\cdots(\Psi_{k,w,\neg s_{ij}}^{\ell}+\beta+(N_{i,j,w}^{\ell}-1))}{(\Psi_{k,\cdot,\neg s_{ij}}^{\ell}+\beta V_{\ell})\cdots(\Psi_{k,\cdot,\neg s_{ij}}^{\ell}+\beta V_{\ell}+(N_{i,j}^{\ell}-1))}\label{eq:lambdasegInference}
\end{align}
\end{small}

\noindent Notice how both assumptions are relaxed is this model: the first term of the result (discussed in the Appendix) shown in Eq. \eqref{eq:lambdasegInference}  accounts for the topic dependence between the paired documents, while the second incorporates the segment boundaries.

\section{Experimental Framework}
 \begin{table}[t]\scriptsize\centering\renewcommand{\arraystretch}{1.4}
  \begin{tabular}{l ccc ccc}
\toprule
          & \multicolumn{3}{c}{Full Dataset} & \multicolumn{3}{c}{Topic Modeling Subsets} \\
          \cmidrule(lr){2-4} \cmidrule(lr){5-7}
  Dataset & $D$ & $N$ & $V$ & $D$ & $N$ & $V$ \\
\midrule
   Wiki$^{\text{En}}_{\text{En-Fr}}$ & 937,991 & 259M & 619,056 & 10,000 & 2.55M &33,925 \\
   Wiki$^{\text{Fr}}_{\text{En-Fr}}$ & 937,991 & 159M & 466,423 & 10,000 & 1.64M &26,604 \\
   Wiki$^{\text{En}}_{\text{En-Ge}}$ & 849,955 & 391M & 599,233 &10,000 &2.54M & 33,198\\
   Wiki$^{\text{Ge}}_{\text{En-Ge}}$ & 849,955 & 391M & 894,798 &10,000 & 1.81M & 44,898 \\
   Wiki$^{\text{En}}_{\text{En-It}}$ & 732,416 & 200M & 519,897&10,000 & 2.55M & 33,934 \\
   Wiki$^{\text{It}}_{\text{En-It}}$ & 732,416 & 125M & 360,760 &10,000 & 1.56M & 25,436\\
   Wiki$^{\text{En}}_{\text{En-Es}}$ & 672,094 &  198M & 497,805 &10,000 & 2.82M & 35,156 \\
   Wiki$^{\text{Es}}_{\text{En-Es}}$ & 672,094 & 135M  & 355,677 &10,000 &1.86M & 25,796 \\
   Wiki$^{\text{En}}_{\text{En-Pt}}$ & 540,467 & 160M & 428,293 &10,000 & 2.86M& 34,687\\
   Wiki$^{\text{Pt}}_{\text{En-Pt}}$ & 540,467 &  61M & 222,547 &10,000 & 1.9M& 19,347 \\
\bottomrule
  \end{tabular}
\caption{Data used for evaluating topic coherence (left) and topic modeling (right) purposes. The names signify the language pair and the language that the statistics correspond to. For instance, Wiki$^{\text{en}}_{\text{en-fr}}$ are the English documents of the En-Fr corpus. }\label{tbl:wikipedia_data_stats}
 \end{table}

   \begin{table*}[t]\scriptsize\centering \setlength{\tabcolsep}{3pt}

 \begin{tabular}{cc cc cc Cc cc cc}
 \toprule
  \multicolumn{6}{c}{BiLDA} & \multicolumn{6}{c}{{\segBiLDA}$_s$ } \\
  \cmidrule(lr){1-6} \cmidrule(lr){7-12}
   \multicolumn{2}{c}{\textcolor{blue}{Topic 3 [City]}}& \multicolumn{2}{c}{ \textcolor{red}{ Topic 5 [Sports]} } & \multicolumn{2}{c}{ \textcolor{orange}{ Topic 8 [Art]} } & \multicolumn{2}{c}{\textcolor{blue}{Topic 3 [City]}}& \multicolumn{2}{c}{ \textcolor{red}{ Topic 5 [Sports]} } & \multicolumn{2}{c}{ \textcolor{orange}{ Topic 8 [Art]} } \\
  En & Fr & En & Fr & En & Fr & En & Fr & En & Fr & En & Fr \\
   \cmidrule(lr){1-2}\cmidrule(lr){3-4}\cmidrule(lr){5-6}\cmidrule(lr){7-8}\cmidrule(lr){9-10}\cmidrule(lr){11-12}

  city  & commun    & team    &championnat    &music      &group   &citi & vill & team & championnat & film & the \\
  popul & situ      &play     &club           &album      &album   &popul & situ & play & club & releas & of     \\
  town  & vill      &first    &premi          &releas     &titre   &area & commun & first & premi & also & film   \\
  area  & référent  &game     &coup           &song       &the     &town & part & world & match & album & and  \\
  locat & région    &player   &match          &record     &and     & locat & grand & leagu & coup & first & sort  \\
  \midrule\midrule
  \multicolumn{6}{c}{$\lambda$-BiLDA} & \multicolumn{6}{c}{$\lambda$-{\segBiLDA}$_s$ }\\
  \cmidrule(lr){1-6} \cmidrule(lr){7-12}\\
   \multicolumn{2}{c}{\textcolor{blue}{Topic 3 [City]}}& \multicolumn{2}{c}{ \textcolor{red}{ Topic 5 [Sports]} } & \multicolumn{2}{c}{ \textcolor{orange}{ Topic 8 [Art]} } & \multicolumn{2}{c}{\textcolor{blue}{Topic 3 [City]}}& \multicolumn{2}{c}{ \textcolor{red}{ Topic 5 [Sports]} } & \multicolumn{2}{c}{ \textcolor{orange}{ Topic 8 [Art]} }\\ 
  En & Fr & En & Fr & En & Fr & En & Fr & En & Fr & En & Fr \\
   \cmidrule(lr){1-2}\cmidrule(lr){3-4}\cmidrule(lr){5-6}\cmidrule(lr){7-8}\cmidrule(lr){9-10}\cmidrule(lr){11-12}

citi     &commun     &team         &championnat     &music   &group & popul & situ & team & championnat & film &  the \\
popul    &vill       &play         &club            &album   &premi& city & espec & play & premi & releas & of\\
town     &situ       &first 	 &premi           &release &sort & also & vill & first & club & album & film\\
refer    &villag     &player       &coup            &song    &the  & area & part & world & coup & also & and\\
area     &référent   &game         &saison          &record  &album & town & grand & leagu & saison & song & sort\\
   
  \bottomrule
 \end{tabular}
\caption{For each of the topic models we present three topics: \textcolor{blue}{City}, \textcolor{red}{Sports},  \textcolor{orange}{Art} for the ``En-Fr'' Wikipedia corpus. 
 Notice the strong intra-semantic (words within a topic) and inter-semantic (topics across languages) coherence.
}\label{tbl:topics_visualized}
\end{table*}

\textbf{The comparable corpora} In order to evaluate the proposed models, we perform a series of evaluation tasks using Wikipedia documents in five language pairs as our comparable corpora. The language pairs are English-French (En-Fr),  English-German (En-Ge), English-Italian (En-It),  English-Spanish (En-Es) and English-Portuguese (En-Pt). Table \ref{tbl:wikipedia_data_stats} shows the basic statistics of these datasets. For topic modeling purposes we have sampled subsets from the full datasets (right part of the table) consisting of 10,000 documents for each pair. Since the sampling was random, it is not the same 10,000 English documents used for every language pair.  Notice in the table that for each pair English is the language with the most words ($N$), which was expected since often Wikipedia lemmas are first written in English and then translated to other languages. This is also why Wikipedia is a suitable comparable corpus;  the English version usually includes more information on a topic compared to other languages. To extract comparable Wikipedia documents one can use the inter-language links.  For the sake of reproducibility, we have used the bilingual corpora as made available by \textit{linguatools}.\footnote{url{http://linguatools.org/tools/corpora/wikipedia-comparable-corpora/}} We have cleaned the documents to remove {\it html} tags and tables using Python v2.7 and  Beautiful Soup v4.5.1.\footnote{https://www.crummy.com/software/BeautifulSoup/bs4/doc/} 
The statistics of Table \ref{tbl:wikipedia_data_stats} are after the pre-processing steps, that include lower-casing, filtering the numerical terms out, stemming using the WordNet stemmer as implemented in \cite{bird2009natural}, stop-word removal using the stopword lists of \cite{bird2009natural} and finally filtering vocabulary terms with less than 4 occurrences in the corpus.

\textbf{The models}  We evaluate the following models for each language pair: (i) {\muLDA} that has been proposed in \cite{mimno2009polylingual}  (ii) {\segBiLDA}$_s$  that was presented above and uses sentences as coherent segments, (iii) {\segBiLDA}$_{b}$ that is {\segBiLDA} with the 1,000\footnote{We use 1K bigrams following the work of \protect\cite{lau2013collocations} who found this number to be the optimal choice for similar corpora.} most frequent bigrams  considered as coherent segments.
Complementary to those models that either build on the bag-of-words assumption or incorporate parts of text structure, we also evaluate 
$\lambda$-{\muLDA}, $\lambda$-{\segBiLDA}$_{s}$ and  $\lambda$-{\segBiLDA}$_{b}$ that extend the first three models for comparable corpora. We have implemented each of these models using Python, Numpy and Scipy. As commonly done, we follow previous work e.g., \cite{blei2003latent}, and we set for each model the Dirichlet hyper-parameters $\alpha=1/K$ and $\beta=0.01$, where $K$ is the number of topics.

\textbf{Training Bilbowa.} To estimate the word representations of the En-Fr, En-Ge, En-Es, En-It and En-Pt pairs we used Bilbowa  \cite{GouwsBC15} to generate a dictionary of word embeddings, where words from two different languages are projected to the same space. We have used the open implementation of Bilbowa\footnote{\url{https://github.com/gouwsmeister/bilbowa}} with its default parameters and the training epochs set to 10.\footnote{-size 100 -window 5 -sample 1e-4 -negative 5 -binary 0 -adagrad 1 -xling-lambda 1 -threads 12 -epochs 10} The model requires parallel text, and for this purpose we used the Europarl corpus \cite{koehn2005europarl}. The statistics of the pairs of languages  for the Europarl data are shown in Table \ref{tbl:europarl_data}.

\begin{table}\scriptsize\centering
 \begin{tabular}{l c cc cc}
 \toprule
 $\ell_1-\ell_2$ & $D$ & $N^{\ell_1}$ & $V^{\ell_1}$ &  $N^{\ell_2}$ & $V^{\ell_2}$ \\\midrule
  En-Fr & 1.92M & 56M & 88,774 & 64M & 130,146 \\
  En-Ge & 1,92M & 53M & 86,691 & 50M & 332,285 \\
  En-Es & 1,96M & 54M & 87,916 & 56M & 161,995 \\
  En-It & 1,90M & 55M & 88,172 & 55M& 155,715\\
  En-Pt & 1.96M & 54M & 88,241 & 56M & 145,112\\
\bottomrule
 \end{tabular}
\caption{Statistics for the Europarl corpus. We used the Europarl data to train the BilBowa word representations. }\label{tbl:europarl_data}
\end{table}

\textbf{Visualizing the topics.} As an initial qualitative evaluation of the learned topics, 
Figure \ref{tbl:topics_visualized} presents for 3 topics (\texttt{City, Sports, Art}) the five words with the highest probability for each of the six topic models. The topics were identified after training each model for 200 Gibbs sampling iterations on the Wiki$_{\text{En-Fr}}$ corpus with $K=10$ topics. 
Visual inspection of the topics reveals that the models produce topics that are intra-semantically coherent, that is the words that constitute the topics are semantically relevant.
Further, the topics are are inter-semantically coherent, that is the topics and aligned across languages and closely related words represent them. For instance, the ``Sports'' topic in English contains mostly word like ``team'', ``play'', ``season'' while in French one can find their (stemmed) translations: ``équipe'', ``jouer'', ``saison''. 
Although reassuring the visual inspection of the topics in not sufficient to compare the models. In the rest, we evaluate the models intrinsically,  that is independently of an application as well as extrinsically in the framework of a cross-lingual document retrieval application.

\subsection{Intrinsic Evaluation of the Topic Models}
\textbf{Normalized PMI} Automatically evaluating the coherence of the topics produced by topic models is a task that has received a lot of attention.
The goal is to measure how coherent or interpretable the produced topics are \cite{MimnoWTLM11}. 
It has been recently found that scoring the topics using  co-occurrence measures, such as the pointwise mutual information (PMI) between the top words of a topic, correlates well with human judgments \cite{NewmanLGB10}. To achieve that, an external corpus like Wikipedia is treated like a meta-document, which is used as the basis to calculate the PMI scores of words using a sliding window and applying the equation:
\[
 PMI(w_i, w_j)=\log \frac{P(w_i, w_j)}{P(w_i)P(w_j)}
\]

Evaluating the topic coherence requires selecting the top-$N$ words of a topic and performing the manual or automatic evaluation. Here, $N$ is a hyper-parameter to be chosen and its value can impact the results. Very recently, 
Lau and Baldwin \cite{lau-baldwin2016} showed that $N$ actually impacts the quality of the obtained results and in particular the correlation with human judgments. In their study they conclude that aggregating the topic coherence scores over several
topic cardinalities, leads to a substantially more stable and robust evaluation.

Following these results, we present in Table \ref{tbl:topic_coherence_scores} the topic coherence scores as measured by the normalized pointwise mutual information (nPMI). 
The scores of nPMI range in [-1,1], where  in the limit of -1 two words $w_1$ and $w_2$ never occur together, while in the limit of +1 they always occur together (complete co-occurrence).
As in \cite{lau-baldwin2016}, for each topic, we aggregate the topic coherence scores over three different topic cardinalities: $N\in\{5,10,15\}$. The reference corpora for calculating the topic coherence for each language are the ``Full Datasets'' of Table \ref{tbl:wikipedia_data_stats} excluding the ``Topic Modeling Subsets''. For English we opt for Wiki$^{\text{En}}_{\text{En-Fr}}$, which is the biggest, whereas for the rest of the languages we use their respective Wikipedia datasets.

In Table \ref{tbl:topic_coherence_scores}, note that in most cases $\lambda$-{\segBiLDA}$_b$ outperforms the rest of the models, while  {\segBiLDA}$_b$ and {\segBiLDA}$_s$ follow.   Notice, how 
{\muLDA} although competitive for low values of $K$ does not perform as well. This is probably due to the fact that the concept of context is encapsulated in the calculation of the nPMI scores, and the {\segBiLDA} topic models explicitly account for this. 
In general, increasing the number of topics from 10 to 25 or 50 seems to improve the performance measured by nPMI. For instance, in the lower part of the Table with the averages across languages, increasing the topics increases the best performance from .135 to .151.  
From the table, it is evident that adapting  the topic models for comparable corpora improves  the scores, apart from the case of  {\segBiLDA}$_s$. For the rest of the models ({\muLDA} and {\segBiLDA}$_b$)  the $\lambda$- counterparts perform better according to the columnwise comparison of the averaged results in the lowest rows of the table.

Although well-correlated with human judgments, for nPMI  we only used a small part of the output of topic models, that is for each topic the top-$N$ words. Furthermore, the evaluation of nPMI suffers in that is does not account for the topical overlap between the learned topics as well as recall gaps within a topic, i.e. lack of terms which should have been ideally included. Therefore, we also report the perplexity scores of held-out documents, whose calculation requires more information from the topic models. 

\begin{table}\scriptsize\centering\setlength{\tabcolsep}{2pt}
 \begin{tabular}{ll cc cc cc}
\toprule
$\ell_2$ & $K$ & {\BiLDA}  &  $\lambda$-{\BiLDA} & {\segBiLDA}$_s$ & $\lambda$-{\segBiLDA}$_s$ & {\segBiLDA}$_b$ & $\lambda$-{\segBiLDA}$_b$ \\
\midrule

En & 10 & $.105_{ \pm .07}$ & $.102_{ \pm .07}$ & $.090_{ \pm .06}$ & $.080_{ \pm .07}$ & $.113_{ \pm .05}$ & $.124_{ \pm .06}$ \\
En & 25 & $.124_{ \pm .10}$ & $.125_{ \pm .04}$ & $.129_{ \pm .11}$ & $.111_{ \pm .07}$ & $.140_{ \pm .08}$ & $.150_{ \pm .07}$ \\
En & 50 & $.132_{ \pm .05}$ & $.129_{ \pm .10}$ & $.125_{ \pm .06}$ & $.125_{ \pm .10}$ & $.157_{ \pm .08}$ & $.155_{ \pm .05}$ \\
\midrule
Fr & 10 & $.114_{ \pm .05}$ & $.114_{ \pm .06}$ & $.105_{ \pm .06}$ & $.053_{ \pm .05}$ & $.088_{ \pm .06}$ & $.125_{ \pm .07}$ \\
Fr & 25 & $.122_{ \pm .06}$ & $.121_{ \pm .06}$ & $.124_{ \pm .08}$ & $.068_{ \pm .03}$ & $.120_{ \pm .07}$ & $.114_{ \pm .05}$ \\
Fr & 50 & $.124_{ \pm .06}$ & $.120_{ \pm .07}$ & $.136_{ \pm .07}$ & $.080_{ \pm .06}$ & $.123_{ \pm .07}$ & $.133_{ \pm .06}$ \\
\midrule
Ge & 10 & $.198_{ \pm .09}$ & $.198_{ \pm .11}$ & $.234_{ \pm .10}$ & $.250_{ \pm .08}$ & $.203_{ \pm .10}$ & $.215_{ \pm .10}$ \\
Ge & 25 & $.174_{ \pm .02}$ & $.173_{ \pm .11}$ & $.235_{ \pm .08}$ & $.235_{ \pm .11}$ & $.187_{ \pm .08}$ & $.176_{ \pm .04}$ \\
Ge & 50 & $.183_{ \pm .03}$ & $.180_{ \pm .02}$ & $.230_{ \pm .09}$ & $.255_{ \pm .10}$ & $.181_{ \pm .05}$ & $.183_{ \pm .04}$ \\
\midrule
It & 10 & $.096_{ \pm .06}$ & $.101_{ \pm .05}$ & $.102_{ \pm .07}$ & $.084_{ \pm .06}$ & $.119_{ \pm .06}$ & $.113_{ \pm .05}$ \\
It & 25 & $.109_{ \pm .04}$ & $.118_{ \pm .05}$ & $.104_{ \pm .09}$ & $.099_{ \pm .02}$ & $.143_{ \pm .07}$ & $.127_{ \pm .06}$ \\
It & 50 & $.119_{ \pm .08}$ & $.125_{ \pm .05}$ & $.122_{ \pm .09}$ & $.131_{ \pm .07}$ & $.137_{ \pm .04}$ & $.142_{ \pm .08}$ \\
\midrule
Es & 10 & $.093_{ \pm .11}$ & $.079_{ \pm .09}$ & $.132_{ \pm .15}$ & $.099_{ \pm .11}$ & $.112_{ \pm .04}$ & $.112_{ \pm .11}$ \\
Es & 25 & $.105_{ \pm .06}$ & $.120_{ \pm .05}$ & $.091_{ \pm .08}$ & $.106_{ \pm .12}$ & $.136_{ \pm .08}$ & $.153_{ \pm .06}$ \\
Es & 50 & $.109_{ \pm .04}$ & $.133_{ \pm .07}$ & $.112_{ \pm .10}$ & $.109_{ \pm .09}$ & $.154_{ \pm .05}$ & $.150_{ \pm .06}$ \\
\midrule
Pt & 10 & $.099_{ \pm .05}$ & $.115_{ \pm .07}$ & $.108_{ \pm .13}$ & $.093_{ \pm .09}$ & $.098_{ \pm .04}$ & $.123_{ \pm .06}$ \\
Pt & 25 & $.129_{ \pm .12}$ & $.120_{ \pm .11}$ & $.164_{ \pm .18}$ & $.145_{ \pm .11}$ & $.131_{ \pm .06}$ & $.124_{ \pm .09}$ \\
Pt & 50 & $.120_{ \pm .10}$ & $.137_{ \pm .07}$ & $.143_{ \pm .15}$ & $.125_{ \pm .06}$ & $.141_{ \pm .08}$ & $.145_{ \pm .10}$ \\
\midrule
 \texttt{avg} & 10 & $.117_{ \pm .07}$ & $.118_{ \pm .08}$ & $.129_{ \pm .10}$ & $.110_{ \pm .08}$ & $.122_{ \pm .06}$ & $\boldsymbol{.135}_{ \pm .07}$ \\
\texttt{avg} & 25 & $.127_{ \pm .07}$ & $.129_{ \pm .07}$ & $.141_{ \pm .11}$ & $.127_{ \pm .08}$ & $\boldsymbol{.143}_{ \pm .07}$ & $.141_{ \pm .06}$ \\
\texttt{avg} & 50 & $.131_{ \pm .06}$ & $.137_{ \pm .07}$ & $.145_{ \pm .09}$ & $.137_{ \pm .08}$ & $.149_{ \pm .06}$ & $\boldsymbol{.151}_{ \pm .06}$ \\
\bottomrule
\end{tabular}
\caption{Topic coherence measured by the nPMI  for each of the models. The averages are calculated for each model and $K$ across languages. Overall,  $\lambda$-{\segBiLDA}$_b$ performs the best. }\label{tbl:topic_coherence_scores}
\end{table}

\textbf{Perplexity}
We continue our evaluation by presenting the results of hold out perplexity scores, calculated on the topic model datasets of Table \ref{tbl:wikipedia_data_stats}. 
Achieving lower perplexity score means that the model can explain unseen data more efficiently, thus it generalizes better and it is, in turn, a better model. As a results, a good model with low perplexity should be able to infer better representations for the unseen documents. As perplexity does not use any real application to evaluate the topic models it is also an intrinsic metric.
For a set of test documents $C$ consisting of $N$ words $\{w_1, \ldots, w_N\}$ the perplexity is calculated using:
\begin{small}
\begin{equation}
 perpl(C) = \exp \left( -\frac{\sum\limits_{i=1}^N \log P(w_i) } {N} \right)\label{eq:perplexity}
\end{equation}
\end{small}
In order to estimate perplexity, the topic distributions of the unseen documents are required. They are obtained by iteratively updating the topic assignments to the words of the unseen (held-out) documents while keeping constant the per-word topic distributions ($\Psi^{\ell}$) learned during training. Here, for the perplexity calculations we assume that the held-out documents form thematically-aligned pairs (as during training) and, depending on the topic model, shared or language-dependent per-document distributions are inferred that are used at Eq. \eqref{eq:perplexity}. In the next section, where we will compare the performance of the models in an extrinsic task, we will ignore the links within the pairs to demonstrate than our models perform well under both settings. 

Table \ref{tbl:perplexity_scores_overview} presents the perplexity scores achieved by the topic models. The reported scores are the averages of 10-fold cross-validation as follows: (i) we split the datasets in 10 disjoint sets, (ii) we repeat the training and perplexity calculation steps 10 times, each time considering the $i$-th set to be the held-out documents and the remaining 9 sets for training. The goal is to exclude any bias due to the split. We present the scores for $K\in\{ 25, 50, 100, 150\}$. In terms of perplexity, $\lambda$-{\segBiLDA}$_b$ and $\lambda$-{\BiLDA}    clearly outperform the rest of the systems consistently for each language and language pair. The third best performing system is {\segBiLDA}$_b$ and {\muLDA} follows.  $\lambda$-{\segBiLDA}$_s$ and {\segBiLDA}$_s$ achieve the worst perplexity scores for every experiment. {\segBiLDA}$_s$, while  competitive when evaluated using the nPMI scores, performs poorly in this task. 

Shown from a different angle, it seems that the systems who build on the bag-of-words assumption ({\muLDA} and $\lambda$-{\muLDA}) consistently outperform those that incorporate the boundaries of large spans like sentences ({\segBiLDA}$_s$ and $\lambda$-{\segBiLDA}$_s$). That was expected as it is line with previous work \cite{balikas2016on}, where incorporating text structure in the form of sentence boundaries was found to lead to higher (worse) perplexity. One the other  hand, incorporating the boundaries of smaller spans like bigrams, helps perplexity performance as $\lambda$-{\segBiLDA}$_b$ seems to be the best performing model overall, especially when the number of topics increases. This is also in line with previous work: \cite{lau2013collocations} showed how bigram boundaries improve the topic model results. In fact, $\lambda$-{\segBiLDA}$_b$ further improves {\segBiLDA}$_b$ who is inspired by \cite{lau2013collocations} since it consistently achieves better perplexity scores. 

Another interesting remark concerns the effect of $\lambda$, whose goal is to adapt the topic models for comparable corpora. Notice that  $\lambda$-{\muLDA}, $\lambda$-{\segBiLDA}$_s$ and $\lambda$-{\segBiLDA}$_b$ outperform {\muLDA}, {\segBiLDA}$_s$ and  {\segBiLDA}$_b$ respectively for each of the experiments and topic values. This highlights the positive effect of the proposed binding mechanism on the achieved perplexity scores. What is more, that was achieved by using a simple yet powerful mechanism (aggregation of word embeddings) for calculating the value of $\lambda$ for each document pair and these results can be potentially refined when applying more complex strategies. Effectively, this is the answer to the question $(Q2)$ that the paper investigates. Adapting topic models for comparable corpora improves their generalization performance and, importantly, these improvements are consistent across different topic models (here {\muLDA}, {\segBiLDA}$_s$ and  {\segBiLDA}$_b$) and different pairs of languages.

\begin{sidewaystable}
\tiny
 \begin{tabular}{ll cc cc cc cc cc cc }
 \toprule
&& \multicolumn{6}{c}{English} & \multicolumn{6}{c}{$\ell_2$} \\
$\ell_2$ & $K$  &  \scriptsize {\muLDA} &$\lambda$-{\muLDA} & {\segBiLDA}$_b$ & $\lambda$-{\segBiLDA}$_b$ & {\segBiLDA}$_s$ & $\lambda$-{\segBiLDA}$_s$ 
&{\muLDA} &$\lambda$-{\muLDA} & {\segBiLDA}$_b$ & $\lambda$-{\segBiLDA}$_b$ & {\segBiLDA}$_s$ & $\lambda$-{\segBiLDA}$_s$     \\
\cmidrule(rl){1-2} \cmidrule(rl){3-8}\cmidrule(rl){9-14}
Fr & 25 & $3423_{\pm 123}$ & $\boldsymbol{3391}_{\pm 113}$ & $3445_{\pm 115}$ & $3383_{\pm 98}$ & $3780_{\pm 234}$ & $3727_{\pm 327}$ & $2709_{\pm 70}$ & $2633_{\pm 68}$ & $2724_{\pm 55}$ & $\textbf{2617}_{\pm 89}$ & $3111_{\pm 158}$ & $2929_{\pm 135}$ \\
Fr & 50 & $3009_{\pm 109}$ & $2957_{\pm 114}$ & $3002_{\pm 86}$ & $\boldsymbol{2944}_{\pm 112}$ & $3460_{\pm 263}$ & $3420_{\pm 341}$ & $2424_{\pm 56}$ & $2320_{\pm 64}$ & $2417_{\pm 58}$ & $\boldsymbol{2312}_{\pm 72}$ & $2891_{\pm 145}$ & $2715_{\pm 100}$ \\
Fr & 100 & $2725_{\pm 120}$ & $\boldsymbol{2634}_{\pm 110}$ & $2685_{\pm 98}$ & $2636_{\pm 87}$ & $3288_{\pm 339}$ & $3236_{\pm 357}$ & $2245_{\pm 43}$ & $2092_{\pm 65}$ & $2194_{\pm 49}$ & $\boldsymbol{2085}_{\pm 80}$ & $2720_{\pm 147}$ & $2598_{\pm 113}$ \\
Fr & 150 & $2642_{\pm 121}$ & $\boldsymbol{2526}_{\pm 109}$ & $2569_{\pm 103}$ & $2527_{\pm 102}$ & $3225_{\pm 309}$ & $3176_{\pm 353}$ & $2197_{\pm 44}$ & $2035_{\pm 53}$ & $2135_{\pm 51}$ & $\boldsymbol{2028}_{\pm 70}$ & $2696_{\pm 126}$ & $2558_{\pm 112}$ \\

\midrule
Ge & 25 & $3338_{\pm 114}$ & $3317_{\pm 83}$ & $3350_{\pm 86}$ & $\boldsymbol{3292}_{\pm 90}$ & $3711_{\pm 171}$ & $3639_{\pm 256}$ & $4532_{\pm 434}$ & $4419_{\pm 426}$ & $4508_{\pm 437}$ & $\boldsymbol{4386}_{\pm 430}$ & $5248_{\pm 505}$ & $4929_{\pm 620}$ \\
Ge & 50 & $2920_{\pm 106}$ & $2873_{\pm 102}$ & $2934_{\pm 67}$ & $\boldsymbol{2859}_{\pm 84}$ & $3379_{\pm 221}$ & $3303_{\pm 277}$ & $3952_{\pm 367}$ & $3791_{\pm 376}$ & $3951_{\pm 374}$ & $\boldsymbol{3772}_{\pm 361}$ & $4727_{\pm 505}$ & $4463_{\pm 542}$ \\
Ge & 100 & $2666_{\pm 115}$ & $2589_{\pm 91}$ & $2625_{\pm 112}$ & $\boldsymbol{2570}_{\pm 116}$ & $3200_{\pm 308}$ & $3152_{\pm 286}$ & $3617_{\pm 341}$ & $\boldsymbol{3408}_{\pm 312}$ & $3577_{\pm 339}$ & $3412_{\pm 310}$ & $4424_{\pm 561}$ & $4227_{\pm 555}$ \\
Ge & 150 & $2581_{\pm 108}$ & $2481_{\pm 105}$ & $2528_{\pm 117}$ & $\boldsymbol{2471}_{\pm 107}$ & $3126_{\pm 290}$ & $3109_{\pm 287}$ & $3554_{\pm 330}$ & $\boldsymbol{3284}_{\pm 303}$ & $3484_{\pm 308}$ & $3314_{\pm 298}$ & $4322_{\pm 558}$ & $4163_{\pm 534}$ \\
\midrule
It & 25 & $3393_{\pm 136}$ & $3364_{\pm 117}$ & $3411_{\pm 139}$ & $\boldsymbol{3360}_{\pm 108}$ & $3780_{\pm 180}$ & $3659_{\pm 195}$ & $2688_{\pm 137}$ & $2606_{\pm 110}$ & $2696_{\pm 152}$ & $\boldsymbol{2589}_{\pm 116}$ & $3140_{\pm 258}$ & $2886_{\pm 233}$ \\
It & 50 & $2994_{\pm 101}$ & $2938_{\pm 100}$ & $2983_{\pm 98}$ & $\boldsymbol{2933}_{\pm 101}$ & $3463_{\pm 167}$ & $3346_{\pm 190}$ & $2405_{\pm 112}$ & $2304_{\pm 85}$ & $2404_{\pm 103}$ & $\boldsymbol{2292}_{\pm 88}$ & $2912_{\pm 215}$ & $2678_{\pm 247}$ \\
It & 100 & $2714_{\pm 94}$ & $2639_{\pm 88}$ & $2691_{\pm 81}$ & $\boldsymbol{2637}_{\pm 87}$ & $3261_{\pm 207}$ & $3147_{\pm 203}$ & $2225_{\pm 108}$ & $2099_{\pm 78}$ & $2210_{\pm 96}$ & $\boldsymbol{2092}_{\pm 72}$ & $2787_{\pm 277}$ & $2561_{\pm 246}$ \\
It & 150 & $2628_{\pm 86}$ & $2535_{\pm 84}$ & $2579_{\pm 88}$ & $\boldsymbol{2531}_{\pm 77}$ & $3208_{\pm 225}$ & $3090_{\pm 203}$ & $2188_{\pm 113}$ & $2036_{\pm 80}$ & $2143_{\pm 102}$ & $\boldsymbol{2030}_{\pm 71}$ & $2730_{\pm 273}$ & $2527_{\pm 259}$ \\
\midrule

Es & 25 & $3178_{\pm 132}$ & $\boldsymbol{3140}_{\pm 127}$ & $3192_{\pm 125}$ & $3147_{\pm 125}$ & $3419_{\pm 166}$ & $3386_{\pm 201}$ & $2367_{\pm 121}$ & $2297_{\pm 118}$ & $2376_{\pm 122}$ & $\boldsymbol{2294}_{\pm 124}$ & $2576_{\pm 133}$ & $2439_{\pm 124}$ \\
Es & 50 & $2800_{\pm 104}$ & $2760_{\pm 113}$ & $2802_{\pm 115}$ & $\boldsymbol{2758}_{\pm 107}$ & $3162_{\pm 169}$ & $3082_{\pm 184}$ & $2128_{\pm 104}$ & $2052_{\pm 108}$ & $2127_{\pm 105}$ & $\boldsymbol{2045}_{\pm 97}$ & $2435_{\pm 126}$ & $2256_{\pm 119}$ \\
Es & 100 & $2540_{\pm 114}$ & $2474_{\pm 104}$ & $2516_{\pm 105}$ & $\boldsymbol{2470}_{\pm 96}$ & $2953_{\pm 157}$ & $2903_{\pm 185}$ & $1974_{\pm 97}$ & $1873_{\pm 92}$ & $1955_{\pm 99}$ & $\boldsymbol{1868}_{\pm 90}$ & $2297_{\pm 90}$ & $2159_{\pm 113}$ \\
Es & 150 & $2449_{\pm 106}$ & $\boldsymbol{2363}_{\pm 96}$ & $2406_{\pm 91}$ & $2367_{\pm 93}$ & $2900_{\pm 166}$ & $2848_{\pm 186}$ & $1928_{\pm 91}$ & $\boldsymbol{1814}_{\pm 88}$ & $1894_{\pm 85}$ & $1815_{\pm 85}$ & $2262_{\pm 105}$ & $2126_{\pm 109}$ \\
\midrule

Pt & 25 & $3219_{\pm 173}$ & $3187_{\pm 177}$ & $3218_{\pm 161}$ & $\boldsymbol{3185}_{\pm 152}$ & $3472_{\pm 332}$ & $3459_{\pm 419}$ & $2139_{\pm 120}$ & $2042_{\pm 101}$ & $2139_{\pm 108}$ & $\boldsymbol{2040}_{\pm 81}$ & $2497_{\pm 180}$ & $2241_{\pm 129}$ \\
Pt & 50 & $2837_{\pm 175}$ & $2812_{\pm 170}$ & $2832_{\pm 157}$ & $\boldsymbol{2811}_{\pm 165}$ & $3201_{\pm 364}$ & $3152_{\pm 419}$ & $1917_{\pm 110}$ & $1809_{\pm 87}$ & $1914_{\pm 99}$ & $\boldsymbol{1797}_{\pm 87}$ & $2337_{\pm 174}$ & $2045_{\pm 121}$ \\
Pt & 100 & $2591_{\pm 180}$ & $2529_{\pm 167}$ & $2555_{\pm 161}$ & $\boldsymbol{2524}_{\pm 166}$ & $2998_{\pm 403}$ & $2980_{\pm 416}$ & $1775_{\pm 104}$ & $1638_{\pm 81}$ & $1752_{\pm 98}$ & $\boldsymbol{1636}_{\pm 78}$ & $2200_{\pm 150}$ & $1945_{\pm 136}$ \\
Pt & 150 & $2506_{\pm 183}$ & $2424_{\pm 170}$ & $2448_{\pm 165}$ & $\boldsymbol{2422}_{\pm 166}$ & $2948_{\pm 383}$ & $2921_{\pm 419}$ & $1739_{\pm 101}$ & $\boldsymbol{1587}_{\pm 75}$ & $1699_{\pm 93}$ & $1593_{\pm 72}$ & $2132_{\pm 149}$ & $1918_{\pm 120}$  \\

\bottomrule
 \end{tabular}
\caption{The perplexity scores achieved by the proposed topic models for five bilingual datasets when $K\in\{ 25, 50, 100, 150\}$. The best (lowest) score achieved per language and $k$ is shown in bold. $\lambda$-{\segBiLDA}$_b$ achieves the best perplexity scores in most of the experiments. }\label{tbl:perplexity_scores_overview}
\end{sidewaystable}

\subsection{Extrinsic Evaluation of the Topic Models}
\textbf{Cross-lingual Document Retrieval}
We conclude the evaluation of the presented topic models by reporting their performance in the framework of a cross-lingual document retrieval application. As discussed during  perplexity evaluation, the model can infer the per-document topic distribution for previously unseen data. Recall, that as Figure \ref{tbl:topics_visualized} depicts, the learned topics are aligned. Therefore, one may perform inference with a trained model in each language separately, without requiring the explicit links between the documents of a  pair. To achieve that, the per-words topic distributions of each language are used. Then, documents with similar topic distributions written in different languages are actually similar due to 
the inter-semantic coherence of the topics alignments between the learned topics. This is a central observation that enables various cross-lingual applications \cite{vulic2015probabilistic} as well as cross-lingual document retrieval. 

The task we propose is a cross-lingual document discovery task (CLDD), The goal is to identify counterpart Wikipedia documents due to cross-language links. In particular, given a document $d_i^{\ell_1}$ as a query, one needs to identify the corresponding document $d_i^{\ell_2}$. For instance, given an English document for ``Dog'' the task is to retrieve the German article for ``Hund'' and, vice versa  given the article for  ``Hund'' one must retrieve the article for ``Dog''.  

Following \cite{fukumasu2012symmetric,vulic2013cross} who found bilingual topic models efficient for the task we address it using the following pipeline. For each of the five language pairs, we train topic models on 9,000 document pairs (18,000 documents). For the remaining 2,000 documents (that is 1,000 pairs of documents) we infer their topic distributions using one language at a time. We consider the cross-language links to be our golden standard. Then, given a document $d_i^{\ell_1}$ whose inferred topic distribution is $\theta_i^{\ell_1}$, we rank every document written at $\ell_2$ according to the  KL-divergence (Kullback-Leibler divergence: \cite{kullback1951information}) between $\theta_i^{\ell_1}$ and $\theta_j^{\ell_2}$ and using the golden links evaluate the performance. 
The KL-divergence measures the distance of probability distributions and is a suitable distance measure for our case as the topic distributions are probability distributions.
We repeat the retrieval experiment 10 times by randomly selecting 500 documents (and their counterparts) out of the 1,000 held-out document pairs. As evaluation measure, we report the scores of Mean Reciprocal Rank (MRR) \cite{voorhees1999trec} that accounts for the rank of the true positive documents in the returned ranked list.\footnote{For cases where there is a single golden documents for each query, MRR is equivalent to Mean Average Precision.} The scores of the MRR evaluation measure are given by: 
\[
\text{MRR} = \frac{1}{|D|} \sum\limits_{i=1}^{|D|} \frac{1}{\text{rank}_i}
\]
\noindent where $|D|$ is the number of documents (queries) at each experiment and $\text{rank}_i$ denotes the rank of the true document to be retrieved.
Further, one has MRR $\in[0,1]$ and the higher the score, the higher the rank of the true positive document in the returned list is. 

Table \ref{tbl:mrr_scores_overview} reports the achieved scores for the document representations inferred for each topic model. The scores are the average performance of the 10 experiments accompanied by the standard deviations. In terms of notation, $\ell_1\rightarrow\ell_2$ (e.g., En$\rightarrow$Fr) stands for the experiment where the documents of $\ell_1$ (e.g., English) are used as queries and the documents of $\ell_2$ (e.g., French) are to be retrieved.  The results of the table clearly establish the improvements on the task due the adaptation of the bilingual topic models for comparable corpora. Notice how $\lambda$-{\muLDA}, $\lambda$-{\segBiLDA}$_b$ outperform the rest of the models and especially their counterparts {\muLDA} and {\segBiLDA}$_b$ in most of the experiments. The observed improvements are consistent across the language pairs and number of topics $K\in\{25, 50, 100, 150\}$.   This suggests that quantifying the semantic similarity between the documents of the pairs during training led to discovering better topics, whose performance we evaluated in the CLDD task by trying to identify the links of held-out document pairs.  

It is to be noted that  $\lambda$-{\segBiLDA}$_s$ and {\segBiLDA}$_s$ both perform poorly on the task. We believe that this is due to the fact that assuming large spans like sentences in Wikipedia documents to be thematically coherent results in per-document topic distributions unable to capture fine-grained differences between documents. In turn, such fine-grained differences are necessary for achieving high performance on the CLDD task.   

Finally, our results suggest that incorporating text structure in the form of short text spans (bigrams) and adapting the bilingual topic models for comparable corpora benefits the performance on CLDD.

\begin{table}\scriptsize \centering\setlength{\tabcolsep}{3pt}
 \begin{tabular}{ll cc cc cc  }
 \toprule
 && \multicolumn{6}{c}{MRR}  \\
\cmidrule(lr){3-8} 
$K$  &$\ell_1\rightarrow\ell_2$ &   \muLDA &$\lambda$-{\muLDA} & {\segBiLDA}$_b$ & $\lambda$-{\segBiLDA}$_b$ & {\segBiLDA}$_s$ & $\lambda$-{\segBiLDA}$_s$ \\
\cmidrule(rl){3-4} \cmidrule(rl){5-6} \cmidrule(rl){7-8} 

25	 &En$\rightarrow$Fr	 & 37.0$_{\pm 1.1}$ 	 & \textbf{39.7}$_{\pm 1.2}$ 	 & 36.0$_{\pm 1.0}$ 	 & 37.1$_{\pm 0.9}$ 	 & 14.6$_{\pm 0.6}$ 	 & 7.6$_{\pm 0.4}$ \\
50	 &En$\rightarrow$Fr	 & 43.8$_{\pm 1.3}$ 	 & \textbf{44.6}$_{\pm 1.1}$ 	 & 41.9$_{\pm 1.3}$ 	 & 41.8$_{\pm 1.5}$ 	 & 13.7$_{\pm 0.6}$ 	 & 14.1$_{\pm 0.7}$ \\
100	 &En$\rightarrow$Fr	 & 43.6$_{\pm 1.4}$ 	 & 45.3$_{\pm 1.9}$ 	 & 42.6$_{\pm 1.9}$ 	 & \textbf{47.2}$_{\pm 1.1}$ 	 & 13.4$_{\pm 0.4}$ 	 & 11.4$_{\pm 0.5}$ \\
150	 &En$\rightarrow$Fr	 & 38.5$_{\pm 2.1}$ 	 & 39.2$_{\pm 1.5}$ 	 & 39.3$_{\pm 1.0}$ 	 & \textbf{42.7}$_{\pm 1.5}$ 	 & 18.0$_{\pm 1.1}$ 	 & 13.7$_{\pm 0.9}$ \\\midrule
25	 &Fr$\rightarrow$En	 & 37.8$_{\pm 0.9}$ 	 & \textbf{39.3}$_{\pm 0.9}$ 	 & 36.7$_{\pm 0.6}$ 	 & 37.6$_{\pm 1.0}$ 	 & 14.3$_{\pm 0.6}$ 	 & 7.5$_{\pm 0.4}$ \\
50	 &Fr$\rightarrow$En	 & 44.0$_{\pm 1.1}$ 	 & \textbf{47.1}$_{\pm 1.3}$ 	 & 43.0$_{\pm 1.2}$ 	 & 44.2$_{\pm 1.1}$ 	 & 13.6$_{\pm 0.8}$ 	 & 14.3$_{\pm 0.7}$ \\
100	 &Fr$\rightarrow$En	 & 45.7$_{\pm 1.2}$ 	 & 45.7$_{\pm 0.9}$ 	 & 44.0$_{\pm 1.2}$ 	 & \textbf{47.7}$_{\pm 1.1}$ 	 & 13.6$_{\pm 0.6}$ 	 & 10.4$_{\pm 0.7}$ \\
150	 &Fr$\rightarrow$En	 & 39.5$_{\pm 1.6}$ 	 & 42.7$_{\pm 1.3}$ 	 & 41.4$_{\pm 1.3}$ 	 & \textbf{45.2}$_{\pm 1.2}$ 	 & 19.3$_{\pm 0.9}$ 	 & 13.6$_{\pm 0.6}$ \\\midrule
25	 &En$\rightarrow$Ge	 & 44.1$_{\pm 1.4}$ 	 & 42.4$_{\pm 1.1}$ 	 & 43.2$_{\pm 1.2}$ 	 & \textbf{45.0}$_{\pm 0.8}$ 	 & 12.6$_{\pm 0.7}$ 	 & 18.5$_{\pm 0.6}$ \\
50	 &En$\rightarrow$Ge	 & 51.7$_{\pm 1.4}$ 	 & \textbf{55.7}$_{\pm 1.0}$ 	 & 49.4$_{\pm 1.0}$ 	 & 52.0$_{\pm 1.1}$ 	 & 19.6$_{\pm 0.8}$ 	 & 15.8$_{\pm 1.1}$ \\
100	 &En$\rightarrow$Ge	 & 51.8$_{\pm 1.2}$ 	 & \textbf{54.2}$_{\pm 0.8}$ 	 & 51.4$_{\pm 0.9}$ 	 & 51.7$_{\pm 1.0}$ 	 & 21.1$_{\pm 0.6}$ 	 & 16.0$_{\pm 0.9}$ \\
150	 &En$\rightarrow$Ge	 & 48.1$_{\pm 1.1}$ 	 & 49.9$_{\pm 1.2}$ 	 & 47.3$_{\pm 0.9}$ 	 & \textbf{51.5}$_{\pm 1.3}$ 	 & 21.8$_{\pm 0.8}$ 	 & 21.1$_{\pm 0.5}$ \\\midrule
25	 &Ge$\rightarrow$En	 & \textbf{43.8}$_{\pm 1.5}$ 	 & 42.9$_{\pm 1.5}$ 	 & 42.6$_{\pm 1.3}$ 	 & \textbf{43.8}$_{\pm 1.3}$ 	 & 13.1$_{\pm 0.6}$ 	 & 18.5$_{\pm 0.8}$ \\
50	 &Ge$\rightarrow$En	 & 49.9$_{\pm 1.2}$ 	 & \textbf{53.6}$_{\pm 1.3}$ 	 & 48.2$_{\pm 1.0}$ 	 & 50.9$_{\pm 1.3}$ 	 & 17.9$_{\pm 1.0}$ 	 & 16.8$_{\pm 1.2}$ \\
100	 &Ge$\rightarrow$En	 & 51.5$_{\pm 1.1}$ 	 & \textbf{53.7}$_{\pm 1.1}$ 	 & 50.9$_{\pm 0.8}$ 	 & 52.8$_{\pm 1.2}$ 	 & 20.7$_{\pm 0.9}$ 	 & 16.7$_{\pm 0.9}$ \\
150	 &Ge$\rightarrow$En	 & 46.8$_{\pm 1.3}$ 	 & 46.8$_{\pm 1.0}$ 	 & 46.2$_{\pm 1.3}$ 	 & \textbf{50.4}$_{\pm 1.5}$ 	 & 20.6$_{\pm 0.5}$ 	 & 20.3$_{\pm 1.0}$ \\\midrule
25	 &En$\rightarrow$It	 & \textbf{36.4}$_{\pm 1.2}$ 	 & 34.9$_{\pm 0.8}$ 	 & 33.8$_{\pm 0.5}$ 	 & 34.2$_{\pm 1.3}$ 	 & 9.2$_{\pm 0.6}$ 	 & 6.6$_{\pm 0.6}$ \\
50	 &En$\rightarrow$It	 & 38.9$_{\pm 1.7}$ 	 & 38.3$_{\pm 1.2}$ 	 & 37.4$_{\pm 1.4}$ 	 & \textbf{39.8}$_{\pm 1.0}$ 	 & 13.8$_{\pm 0.7}$ 	 & 10.5$_{\pm 0.4}$ \\
100	 &En$\rightarrow$It	 & 39.0$_{\pm 1.2}$ 	 & 38.9$_{\pm 1.1}$ 	 & 39.0$_{\pm 1.5}$ 	 & \textbf{41.1}$_{\pm 1.4}$ 	 & 14.6$_{\pm 0.5}$ 	 & 12.0$_{\pm 0.6}$ \\
150	 &En$\rightarrow$It	 & 35.9$_{\pm 1.1}$ 	 & 37.1$_{\pm 0.6}$ 	 & \textbf{38.8}$_{\pm 1.4}$ 	 & 37.8$_{\pm 1.0}$ 	 & 13.0$_{\pm 0.5}$ 	 & 10.8$_{\pm 0.6}$ \\\midrule
25	 &It$\rightarrow$En	 & \textbf{35.5}$_{\pm 0.9}$ 	 & 35.3$_{\pm 1.1}$ 	 & 33.7$_{\pm 0.8}$ 	 & 35.2$_{\pm 1.2}$ 	 & 8.8$_{\pm 0.5}$ 	 & 6.4$_{\pm 0.5}$ \\
50	 &It$\rightarrow$En	 & 39.7$_{\pm 1.5}$ 	 & 39.4$_{\pm 1.0}$ 	 & 37.2$_{\pm 1.5}$ 	 & \textbf{40.2}$_{\pm 1.4}$ 	 & 13.2$_{\pm 0.5}$ 	 & 10.8$_{\pm 0.6}$ \\
100	 &It$\rightarrow$En	 & 40.6$_{\pm 1.5}$ 	 & 40.0$_{\pm 1.1}$ 	 & 39.1$_{\pm 1.0}$ 	 & \textbf{40.8}$_{\pm 1.2}$ 	 & 14.8$_{\pm 0.7}$ 	 & 12.6$_{\pm 0.4}$ \\
150	 &It$\rightarrow$En	 & 37.4$_{\pm 1.2}$ 	 & \textbf{39.8}$_{\pm 1.2}$ 	 & 37.4$_{\pm 1.7}$ 	 & 39.3$_{\pm 1.6}$ 	 & 13.4$_{\pm 0.9}$ 	 & 11.2$_{\pm 0.6}$ \\\midrule
25	 &En$\rightarrow$Pt	 & 33.8$_{\pm 1.1}$ 	 & \textbf{34.9}$_{\pm 1.1}$ 	 & 33.6$_{\pm 1.3}$ 	 & 34.3$_{\pm 1.3}$ 	 & 8.8$_{\pm 0.4}$ 	 & 10.2$_{\pm 0.8}$ \\
50	 &En$\rightarrow$Pt	 & 38.2$_{\pm 1.2}$ 	 & 38.3$_{\pm 1.5}$ 	 & 37.7$_{\pm 1.1}$ 	 & \textbf{39.0}$_{\pm 1.3}$ 	 & 14.4$_{\pm 0.8}$ 	 & 11.2$_{\pm 0.4}$ \\
100	 &En$\rightarrow$Pt	 & 38.9$_{\pm 1.3}$ 	 & 38.3$_{\pm 1.1}$ 	 & 38.1$_{\pm 1.0}$ 	 & \textbf{40.2}$_{\pm 1.0}$ 	 & 16.5$_{\pm 0.9}$ 	 & 10.0$_{\pm 0.4}$ \\
150	 &En$\rightarrow$Pt	 & 35.0$_{\pm 1.6}$ 	 & 35.9$_{\pm 1.8}$ 	 & 34.5$_{\pm 1.6}$ 	 & \textbf{36.3}$_{\pm 1.6}$ 	 & 14.8$_{\pm 0.5}$ 	 & 11.2$_{\pm 0.5}$ \\\midrule
25	 &Pt$\rightarrow$En	 & 35.8$_{\pm 1.3}$ 	 & \textbf{36.2}$_{\pm 1.4}$ 	 & 34.4$_{\pm 1.2}$ 	 & 36.0$_{\pm 0.8}$ 	 & 9.7$_{\pm 0.6}$ 	 & 11.2$_{\pm 0.8}$ \\
50	 &Pt$\rightarrow$En	 & 39.5$_{\pm 1.6}$ 	 & 40.0$_{\pm 1.6}$ 	 & 38.2$_{\pm 1.1}$ 	 & \textbf{40.3}$_{\pm 1.3}$ 	 & 15.4$_{\pm 0.8}$ 	 & 12.7$_{\pm 0.5}$ \\
100	 &Pt$\rightarrow$En	 & 41.0$_{\pm 0.9}$ 	 & 40.1$_{\pm 1.4}$ 	 & 40.5$_{\pm 0.9}$ 	 & \textbf{43.4}$_{\pm 0.9}$ 	 & 18.9$_{\pm 0.6}$ 	 & 10.0$_{\pm 0.6}$ \\
150	 &Pt$\rightarrow$En	 & 36.4$_{\pm 1.2}$ 	 & 40.0$_{\pm 1.3}$ 	 & 37.3$_{\pm 1.2}$ 	 & \textbf{41.0}$_{\pm 1.4}$ 	 & 14.6$_{\pm 1.0}$ 	 & 12.6$_{\pm 0.6}$ \\

\bottomrule
 \end{tabular}
\caption{The  scores for the CLDD task achieved by the proposed topic models for five bilingual datasets when $K\in\{ 25, 50, 100, 150\}$. The best (highest) score achieved per language and $k$ is shown in bold. The topic distributions induced by $\lambda$-{\segBiLDA}$_b$ achieved the highest MRR scores in most of the experiments. }\label{tbl:mrr_scores_overview}
\end{table}

\section{Discussion and Conclusion}

In this paper we argued that {\muLDA}, which is the probably the most representative multilingual  topic model, can be extended to be adapted to extract topics from comparable corpora. To this end, we proposed $\lambda$-{\muLDA} which relaxes the assumption of {\muLDA}, where aligned documents shared identical topic distributions. Using recent work on cross-lingual word embeddings we proposed an efficient way to estimate the semantic similarity $\lambda_i$ for the documents of an aligned pair which was then incorporated to the novel topic model.

The concept of $\lambda$-{\muLDA} is general enough and can be applied in a plethora of models. To demonstrate that, we extended in the bilingual setting a class of topic models whose goal is to incorporate text structure in the form of coherent text segment boundaries. We further adapted them for comparable corpora and we derived a Gibbs sampling approach for topic inference. 

In our evaluation framework, we demonstrated the advantages of the proposed topic models. In summary, we found that incorporating text structure and adapting for comparable corpora improved topic coherence measure by nPMI, the generalization performance measured by perplexity and the performance on the CLDD task. Importantly, adapting the models for comparable corpora, consistently improved the generalization performance for each type of topic model we tried, which signifies the applicability of the proposed approach.

The methods and results presented in this work open several avenues for future research. The advantages of incorporating text structure in the form of boundaries of coherent segments motivate further research on the type of the spans to be considered that would achieve the best performance. For instance, one may use linguistically motivated text spans like noun phrases, statistically motivated text spans, like $n$-grams where $n>2$ or even suggest unsupervised methods based on sampling to identify coherent segments that would improve the performance of topic models in the line of \cite{du2010segmented,du2013topic}. 

Furthermore, adapting {\muLDA} for comparable corpora critically relies on the estimation of $\lambda$. Here, to generate document representations we applied the \texttt{average} compositional function on the top of word embeddings learned with Bilbowa. One, may imagine other ways to estimate $\lambda$ and the recent progress  on deep neural networks is expected to provide better solutions. 

Another direction of research concerns the binding mechanism between the topic distributions of each language. Here, we proposed to incorporate the binding in the priors of the \texttt{Dirichlet} distributions. In our future research we aim at investigating different mechanisms such as more elaborate prior distributions to achieve this binding. Further, by having access to cross-lingual word embeddings one may imagine topic models that do not rely pairs of aligned documents in their inputs. Such models should rather a retrieval step to identify or quantify the possible alignments.

\section*{A. Gibbs Sampling Equations for the proposed topic models}
We derive the Gibbs sampling equations for $\lambda$-{\segBiLDA}:
\begin{scriptsize}
\begin{align} 
~&\text{sample } z_{i,j}^{\ell} \sim P\Big( z_{i,j}^{\ell}=k|\boldsymbol{z}_{\neg{s_{i,j}}}^{\ell}, \boldsymbol{z}^{\Othlang}, \boldsymbol{w}^{\ell}, \boldsymbol{w}^{\Othlang},\alpha, \beta, \lambda_i, \theta^{\ell}, \theta^{\Othlang} \Big) \nonumber\\ 
~&\!\!\!\!\propto\!\! \int\limits_{\theta_i^\ell} \!\int\limits_{\phi} P\Big( z_{s_{i,j}}^{\ell}=k|\boldsymbol{z}_{\neg{s_{i,j}}}^{\ell}, \boldsymbol{w}^{\ell},\alpha, \beta, \lambda_i, \theta^{\ell}, \theta^{\Othlang} \Big) d\phi d\theta_i^\ell \nonumber\\
&\!\!\!\!\propto \!\! \int\limits_{\theta_i^\ell} \!\int\limits_{\phi} P(z_{s_{i,j}}^{\ell}=k|\boldsymbol{z}^{\ell}_{\neg {s_{i,j}}}, \theta^{\ell},  \theta^{\Othlang}, \lambda_i, \alpha) \times P(\boldsymbol{w}_{i,j}^{\ell}|z_{i,j}^{\ell}=k, \boldsymbol{z}_{\neg {i,j}}^{\ell},\boldsymbol{w}_{\neg {i,j}}^{\ell},\phi, \beta ) d\phi d\theta_i^\ell \nonumber\\
&\!\!\!\!\propto \!\! P(z_{s_{i,j}}^{\ell}=k|\boldsymbol{z}^{\ell}_{\neg {s_{i,j}}},  \theta^{\Othlang}, \lambda_i, \alpha)\times\left(\int\limits_{\theta_i^\ell} \!\int\limits_{\phi}P(\boldsymbol{w}_{s_{i,j}}^{\ell}|z_{s_{i,j}}^{\ell}=k, \boldsymbol{z}_{\neg {s_{i,j}}}^{\ell},\boldsymbol{w}_{\neg {s_{i,j}}}^{\ell},\phi, \beta ) d\phi d\theta_i^\ell\right)\nonumber
\end{align}
\end{scriptsize}
\noindent For the first term, observe that sampling $P(z_{s_{i,j}}^{\ell}=k|\boldsymbol{z}^{\ell}_{\neg {s_{i,j}}},  \theta^{\Othlang}, \lambda_i, \alpha)$ is exactly the same with sampling $P(z_{s_{i,j}}^{\ell}=k|\boldsymbol{z}^{\ell}_{\neg {i,j}})$ in the case of standard LDA, replacing the Dirichlet parameter $\alpha$ with $\alpha+\lambda_i \theta^{\Othlang}$. 
The derivation of the second term where segments have several words, follows the steps shown on \cite{balikas2016on}. Hence, we deduce that:
\begin{scriptsize}
\begin{align}
 P\Big( z_{i,j}^{\ell}=z_k|\boldsymbol{z}_{\neg s_{i,j}}^{\ell}, \boldsymbol{w}^{\ell},\alpha, \beta, \lambda_i, \theta^{\Othlang} \Big) \propto 
 \frac{\Omega_{d,k,\neg s_{i,j}}^{\ell}+\alpha+\lambda_i\theta_d^{\Othlang}}{\Omega_{d,\cdot,\neg s_{i,j}}^{\ell}  + K\alpha+ K\lambda_i } \times \nonumber\\
\times \frac{\prod\limits_{w\in s_{ij}^{\ell_1}}(\Psi_{k,w,\neg s_{ij}}^{\ell}+\beta)\cdots(\Psi_{k,w,\neg s_{ij}}^{\ell}+\beta+(N_{i,j,w}^{\ell}-1))}{(\Psi_{k,\cdot,\neg s_{ij}}^{\ell}+\beta V_{\ell})\cdots(\Psi_{k,\cdot,\neg s_{ij}}^{\ell}+\beta V_{\ell}+(N_{i,j}^{\ell}-1))} \label{eq:result}
\end{align}
\end{scriptsize}

\noindent In the last result,  for Gibbs sampling the fraction of the first term can be simplified by omitting the denominator as in \cite{heinrich2005parameter,balikas2016on}:

\begin{scriptsize}
\begin{equation}
\frac{\Omega_{d,k,\neg s_{i,j}}^{\ell}+\alpha+\lambda_i\theta_d^{\Othlang}}{\Omega_{d,\cdot,\neg s_{i,j}}^{\ell}  + K\alpha+ K\lambda_i } \sim[\Omega_{d,k,\neg s_{i,j}}^{\ell}+\alpha+\lambda_i\theta_d^{\Othlang}]   \label{eq:simplification}
\end{equation}
\end{scriptsize}

\noindent Integrating Eq. \eqref{eq:simplification} to Eq. \eqref{eq:result}  leads to the desired result.

The equations for $\lambda$-{\muLDA} and {\segBiLDA} are simpler. For $\lambda$-{\muLDA} ones does not have segments and the product in Eq. \eqref{eq:lambdasegInference} and in the subsequent calculations is simplified to a simple term. For {\segBiLDA} there is single topic distribution and hence $\lambda_i=1$, which results in a single $\Omega$ counter matrix, that holds the counts for both documents.

\bibliographystyle{plain}
\bibliography{references}  
%
%


\end{document}